\documentclass[lettersize,journal]{IEEEtran}
\usepackage{amsmath,amsfonts}
\usepackage{algorithmic}
\usepackage{algorithm}
\usepackage{array}
\usepackage[caption=false,font=normalsize,labelfont=sf,textfont=sf]{subfig}
\usepackage{textcomp}
\usepackage{stfloats}
\usepackage{url}
\usepackage{verbatim}
\usepackage{graphicx}
\usepackage{cite}
\usepackage{amssymb}
\usepackage{booktabs}
\usepackage{array}
\usepackage{enumitem}
\usepackage{soul}
\usepackage{multirow}      
\usepackage{threeparttable} 
\usepackage{tabularx}
\usepackage{xcolor}        
\usepackage{amsmath}

\usepackage{pifont}
\newcommand{\cmark}{\ding{51}}%
\newcommand{\xmark}{\ding{55}}%

\begin{document}

\title{CLOSER-VLN: \underline{C}losed-\underline{Lo}op \underline{Se}lf-V\underline{e}rified \underline{R}etrieval-Augmented Reasoning for Aerial Vision-Language Navigation}

\author{Shaoxuan~Li$^{*}$, 
    Xiangyu~Dong$^{*}$, 
    Xiaoguang~Ma$^{\dagger}$, \IEEEmembership{Member, IEEE,} 
    Junfeng~Chen, 
    
    Haoran~Zhao, 
    and Yaoming~Zhou$^{\dagger}$, \IEEEmembership{Member, IEEE}%
    \thanks{$^{*}$Shaoxuan Li and Xiangyu Dong contribute equally to this work.}
    \thanks{$^{\dagger}$Corresponding authors: Xiaoguang Ma and Yaoming Zhou.}
    \thanks{Shaoxuan Li, Xiangyu Dong, and Xiaoguang Ma are with the Foshan Graduate School of Innovation, Northeastern University, Foshan 528311, China (e-mail: 2470864@stu.neu.edu.cn; dxy1999ai@163.com; maxg@mail.neu.edu.cn).}
    \thanks{Xiaoguang Ma is also with the Faculty of Robot Science and Engineering, Northeastern University, Shenyang 110819, China.}
    \thanks{Shaoxuan Li and Xiaoguang Ma are also with the College of Information Science and Engineering, Northeastern University, Shenyang 110819, China.}
    \thanks{Junfeng Chen is with Meituan, Shenzhen, China (e-mail: chenjunfeng@stu.pku.edu.cn).}
    \thanks{Haoran Zhao and Yaoming Zhou are with the School of Aeronautic Science and Engineering, Beihang University, Beijing 100191, China (e-mail: zhaohaoran@buaa.edu.cn; zhouyaoming@buaa.edu.cn).}
    \thanks{Xiaoguang Ma and Haoran Zhao are also with QingniaoAI, China.}
}



\maketitle

\begin{abstract}
Vision-language navigation (VLN) has recently advanced with large language and multimodal models, enabling agents to follow natural-language instructions in unseen environments without training a task-specific navigation policy. However, most existing VLN methods relying on large models still adopt an open-loop decision-execution approach, where candidate actions are generated from instructions and observations but are rarely verified or corrected before execution. This causes critical issues in aerial VLN, where minor errors in intermediate actions may quickly accumulate into large trajectory deviations and lead to target loss. To address this issue, we propose \underline{C}losed-\underline{lo}op \underline{S}elf-v\underline{e}rified \underline{R}etrieval-augmented Reasoning (CLOSER), a training-policy-free method that sequentially performs action reasoning, reliability verification, targeted retrieval, and action correction in a closed-loop manner before executing concrete actions. We instantiate the CLOSER in aerial VLN tasks and develop a CLOSER-VLN framework, which is composed of three components: a hierarchical reasoner for generating candidate actions based on available information, a multidimensional action verifier for assessing the reliability of actions generated by the reasoner, and a verification-triggered multimodal retriever for retrieving targeted exemplars from a memory bank only when verification fails. We conduct experimental evaluations on the CityNav benchmark, where CLOSER-VLN achieves 32.01\% SR and 21.28\% SPL on the test-unseen split, confirming the effectiveness of closed-loop reasoning.

\end{abstract}


\IEEEpubidadjcol
\section{Introduction}

\IEEEPARstart{V}{ision}-language navigation (VLN) aims to enable an embodied agent to reach a target location by following natural-language instructions in previously unseen environments \cite{xu2026dream,an2024etpnav,he2026fine,lin2024correctable}. This task requires the agent to jointly understand language instructions, perceive visual surroundings, reason about spatial relations, and make sequential navigation decisions. In recent years, with the development of Multimodal Large Language Models (MLLMs) and Vision-Language Models (VLMs) \cite{lin2026evolvenav,chen2025constraint,ahn2022can,wang2026imaginenav++,lin2025navcot,cai2025flightgpt}, zero-shot VLN has achieved significant progress, allowing agents to perform instruction-following navigation in unseen environments without training a task-specific navigation policy \cite{zhou2024navgpt,qiao2025open,chen2025constraint}.
Leveraging LLMs as core controllers for reasoning strategies provides a promising route toward building general-purpose embodied navigation agents \cite{ahn2022can,huang2022inner,brohan2023rt2}.

However, most existing zero-shot VLN methods still follow an open-loop decision-execution approach \cite{zhou2024navgpt,chen2024mapgpt,chen2025constraint}. Given a language instruction and the current observation, the agent typically predicts a candidate action and directly executes it. Although some methods introduce maps, memories, or retrieval modules to enhance action prediction credibility \cite{chen2024mapgpt,li2026cmmr,luo2026ragnav}, these methods lack explicit pre-execution action verification. Such an approach is inherently fragile for sequential navigation: a locally unreliable action may not only violate the current instruction or visual evidence, but also change subsequent observation distributions, causing errors to accumulate along the navigation trajectory. In fact, a key limitation of current VLN lies not only in insufficient single-step reasoning ability, but also in the lack of an explicit mechanism to detect and correct unreliable actions before execution.

This issue is particularly pronounced in aerial VLN. Compared with ground-based scenarios, aerial navigation not only involves longer flight distances and a broader operational coverage, but also imposes far more stringent demands on the agent's spatial cognition, long-horizon decision-making, and system reliability \cite{liu2023aerialvln,lee2025citynav}. A small erroneous movement may lead to large trajectory deviations and consequently lead to target loss. Therefore, reliable aerial VLN requires not only stronger action prediction capabilities, but also the ability to examine, diagnose, and correct candidate actions before environment step execution.

Retrieval-Augmented Generation (RAG) provides a feasible way to supplement missing knowledge for navigation methods using large models as core reasoners \cite{borgeaud2022improving}. By retrieving relevant exemplars, historical memories, or external knowledge, an agent can improve semantic understanding and obtain additional decision evidence in unseen environments \cite{monaci2025rana,yoo2024exploratory}. However, existing RAG embodied navigation methods usually use retrieved results as auxiliary context for action generation \cite{luo2026ragnav,li2026cmmr}. Retrieval is often performed before or during reasoning, rather than being explicitly triggered by action verification failure. As a result, conventional RAG can expand the context available to the model, but it cannot directly answer the following key questions: when the agent should question its own generated actions, what knowledge is missing in the current decision, and how actions should be corrected before environment step execution.

\begin{figure*}[t]
\centering
\includegraphics[width=\linewidth]{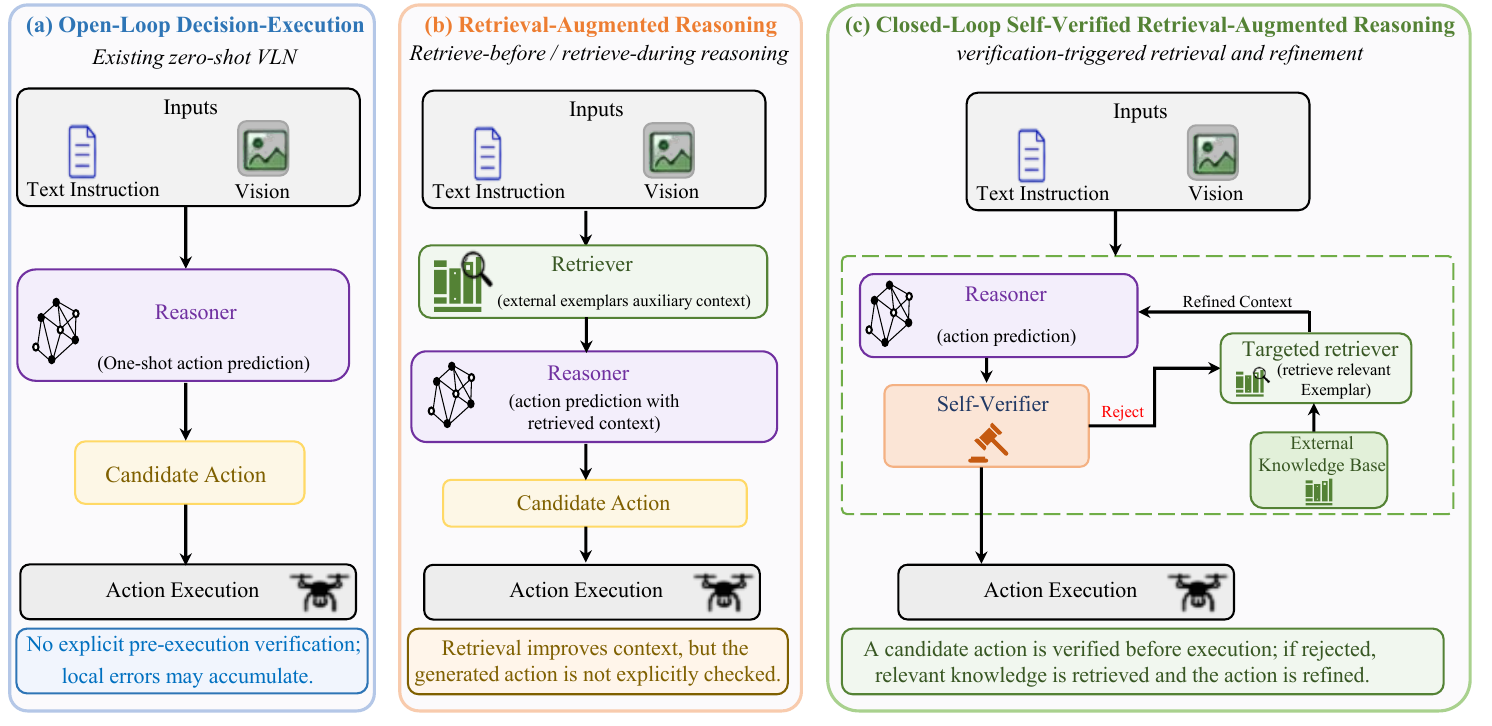}
\caption{Comparison of decision-making methods for aerial vision-language navigation. Zero-shot end-to-end methods (a) and retrieval-augmented methods (b) follow open-loop single-step inference, where the predicted action is directly executed without explicit pre-execution verification. In contrast, the proposed closed-loop self-verified retrieval-augmented reasoning method (c) verifies candidate actions before execution and triggers targeted retrieval only when verification fails, enabling unreliable actions to be corrected before environment step execution.}
\label{fig_1}
\end{figure*}

To address the above issues, we propose \textbf{Closed-loop Self-verified Retrieval-augmented Reasoning (CLOSER)}, a training-policy-free method (Figure~\ref{fig_1}). Instead of directly executing the first predicted action, CLOSER iteratively performs action reasoning, reliability verification, targeted retrieval, and action correction before environment step execution, enabling the system to determine when to question the reliability of its own actions, analyze errors and supplement knowledge when actions are unreliable, and revise the action accordingly. Meanwhile, a maximum number of iterations $K$ within a single step is set to prevent the system from falling into local loops that hinder task progression. This bounded closed-loop process enables unreliable actions to be detected and improved before environment step execution.

Based on the CLOSER method, we further propose \textbf{CLOSER-VLN}, a closed-loop framework specially designed for reliable aerial VLN. CLOSER-VLN follows the standard aerial VLN task interface \cite{xu2026geonav} and organizes each navigation episode into three stages: navigation, search, and localization. To adapt the abstract CLOSER workflow into this specific context, the framework instantiates three specialized components: a hierarchical reasoner responsible for generating candidate actions based on the visual and language information at the current stage; a multidimensional action verifier that serves as a closed-loop gatekeeper—evaluating instruction consistency, visual consistency, and path feasibility during the navigation and search phases while employing a heuristic geometric check to verify the reliability of predicted positions during the localization stage; and a verification-triggered multimodal retriever that, when verification fails, generates a targeted query from the erroneous information to retrieve external exemplars and uses the retrieved results to assist the reasoner in correcting its decision. The maximum number of iterations per step is set to $K=3$.

The main contributions of this paper are summarized as follows:
\begin{itemize}
    \item We propose \textbf{Closed-loop Self-verified Retrieval-augmented Reasoning (CLOSER)}, a closed-loop decision correction method for VLN tasks. The CLOSER introduces reliability verification before action execution and triggers targeted retrieval and action correction when verification fails, thereby alleviating the difficulty of detecting and correcting errors in open-loop action prediction.
    \item We propose \textbf{CLOSER-VLN}, which introduces a bounded closed-loop decision process to improve the reliability of aerial navigation in unseen urban scenarios.
    \item We design a multidimensional action verifier and a verification-triggered multimodal retriever. These two components work collaboratively, allowing the system to identify unreliable candidate actions and perform targeted action correction based on stage-aligned external exemplars.
    \item Comprehensive experiments on the CityNav benchmark show that CLOSER-VLN achieves 32.01\% SR and 21.28\% SPL on the test-unseen split, which represent improvements of 23.6\% and 32.8\% compared to the baselines, validating its effectiveness in aerial VLN.
\end{itemize}

\section{Related Work}
\subsection{Vision-Language Navigation}
Vision-language navigation (VLN) requires an embodied agent to follow natural-language instructions and navigate toward a target in unseen environments. Early VLN studies mainly focus on indoor scenes and learn instruction-following policies through supervised learning, imitation learning, auxiliary progress estimation, cross-modal matching, or trajectory-based pretraining \cite{chen2021topological}. Related benchmarks and embodied navigation settings further extend instruction following to multilingual grounding, remote object localization, continuous environments, and semantic exploration \cite{chaplot2020object}. With the development of LLMs and VLMs, recent studies have begun to explore VLN approaches that employ large models without task-specific training as the core decision-making and control controller, leveraging their powerful multimodal fusion capabilities for instruction decomposition, spatial reasoning, progress estimation, and action prediction \cite{zhou2024navgpt,chen2024mapgpt,qiao2025open,chen2025constraint,dong2025se}. These methods demonstrate that foundation models can support navigation tasks without the need for training a task-specific navigation policy.

Aerial VLN further extends VLN to large-scale outdoor environments with three-dimensional motion and stricter reliability requirements. Aerial VLN introduces UAV-oriented vision-language navigation tasks \cite{liu2023aerialvln}, while CityNav provides a real-world city-scale benchmark with geographic information \cite{lee2025citynav}. Recent aerial VLN methods, such as STMR \cite{gao2024exploring}, CityNavAgent \cite{zhang2025citynavagent}, GeoNav \cite{xu2026geonav}, and ViSA-enhanced Aerial VLN \cite{tong2026visa}, improve UAV navigation through semantic-topo-metric representations, hierarchical semantic planning, global memory, or geospatial reasoning. However, most aerial VLN methods still formulate navigation as open-loop action prediction and generated actions are rarely verified or revised before execution.

\subsection{Retrieval-Augmented Reasoning for Embodied Agents}

RAG introduces external non-parametric memory to supplement the limited knowledge stored in model parameters and has been applied to knowledge-intensive NLP, open-domain question answering, decision support, and scientific synthesis \cite{asai2026synthesizing}. This idea has recently been extended to embodied agents, where retrieval can provide semantic memories, prior observations, task demonstrations, environment knowledge, or navigation experiences for planning and decision-making \cite{mon2025embodied,chen2026retrieval}. For example, Embodied-RAG builds hierarchical embodied memory for retrieval and generation \cite{xie2024embodied}, RANa retrieves prior observations from previous episodes to improve navigation \cite{monaci2025rana}, and ExRAP incorporates retrieval into continual embodied instruction following \cite{yoo2024exploratory}. In VLN, RAGNav and CMMR-VLN further show the potential of retrieval-augmented topological reasoning and continual multimodal memory retrieval for navigation \cite{luo2026ragnav,li2026cmmr}.

These methods demonstrate that external memory can improve grounding and generalization, especially when the agent relies on frozen foundation models. However, retrieval is typically used as auxiliary context for action generation, either before reasoning or as a general memory access mechanism. It is rarely coupled with action verification to determine whether a candidate action should be trusted or revised. Different from conventional retrieval-before-reasoning methods, our work uses retrieval as part of a closed-loop correction process after an unreliable action is detected.

\subsection{Self-Verification}

Self-verification has been widely studied to improve the reliability of LLM reasoning. Reasoning-time prompting methods, including chain-of-thought prompting, self-consistency, ReAct, and Tree of Thoughts, show that intermediate reasoning and deliberate search can improve complex decision-making \cite{wei2022chain,wang2023selfconsistency,yao2023react,yao2023tree}. LLM-as-a-Judge shows that strong language models can serve as evaluators of generated outputs \cite{zheng2023judging}. Iterative self-feedback methods, such as Self-Refine and Reflexion, improve model responses through critique, feedback, and revision \cite{madaan2023self,shinn2023reflexion}. More explicit verification frameworks, including Chain-of-Verification and CRITIC, further demonstrate that deliberate checking and external feedback can improve reasoning quality \cite{dhuliawala2024chain,gou2024critic}. In the multimodal domain, recent studies also extend self-reflective mechanisms to visual reasoning and knowledge-based visual question answering \cite{cocchi2025augmenting}.

Most existing self-verification methods are designed for static text generation, question answering, or vision-language reasoning, where errors mainly affect the final response. Embodied navigation poses a different challenge: an erroneous intermediate action changes the agent's state, affects future observations, and may lead to compounding failures. Therefore, reliable navigation requires action-level verification before execution. Our work extends self-verification to sequential aerial VLN and combines it with targeted retrieval for closed-loop action revision.

\begin{figure*}[t]
\centering
\includegraphics[width=\linewidth]{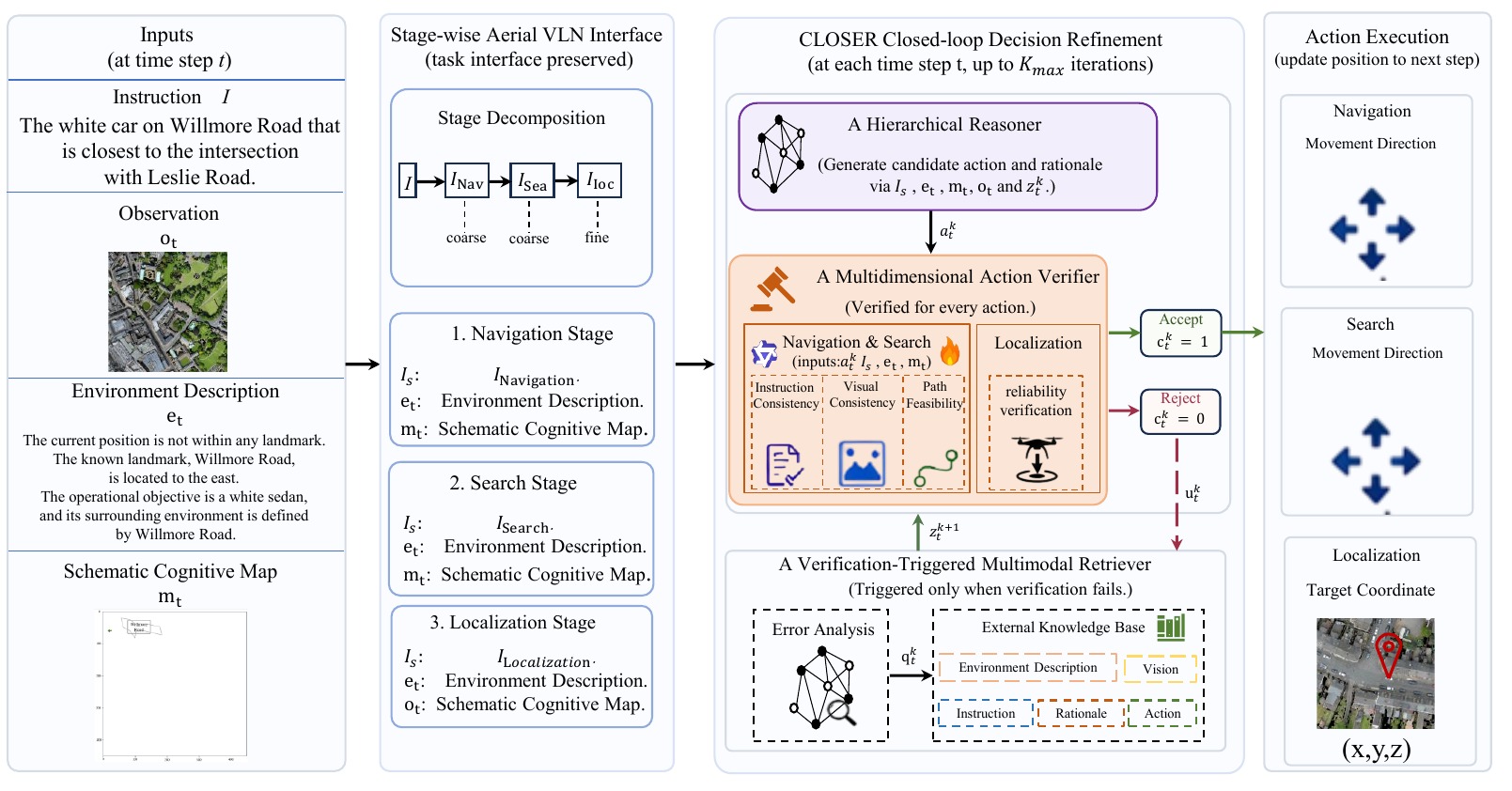}
\caption{Overall architecture of CLOSER-VLN. The CLOSER-VLN preserves the stage-wise aerial VLN interface inherited from GeoNav, while introducing a bounded closed-loop pre-execution refinement process composed of hierarchical reasoning, multidimensional action verification, verification-triggered retrieval, and action refinement.}
\label{fig_2}
\end{figure*}

\section{Methodology}
\subsection{Task Formulation and Baseline Interface}
We study aerial vision-language navigation, where a UAV needs to reach a target location in an unseen urban environment following a natural-language instruction. Unlike coordinate-based navigation, the absolute target position is unavailable during inference. The agent must infer the target by jointly leveraging language descriptions, visual observations, geospatial-semantic cues, and its navigation history.

The CLOSER-VLN adopts the same external task interface as GeoNav~\cite{xu2026geonav}. Each episode consists of three consecutive stages: navigation, search, and localization. Given an original instruction $I$, the system decomposes it into stage-level sub-instructions $l_s$, where $s \in \{\mathrm{Nav}, \mathrm{Sea}, \mathrm{Loc}\}$. Input and output formats are stage-dependent. During navigation and search, the system uses a schematic cognitive map (SCM) $m_t$ (converted from the raw aerial observation $o_t$) as visual input and outputs a discrete movement direction. During localization, it uses the nodes $\mathcal{N}$ of the hierarchical scene graph (HSG) \cite{xu2026geonav} and the raw observation $o_t$ to predict the target coordinate. 

The CLOSER-VLN does not modify the external input-output interface of the CityNav task, nor does it train an end-to-end navigation policy. The use of the hierarchical reasoner, as well as the construction and maintenance of the SCM and HSG, all follow the same settings as GeoNav \cite{xu2026geonav}. Instead, we propose a closed-loop pre-execution decision refinement process in which each candidate action is verified, diagnosed, and, if necessary, supplemented with relevant exemplars before being refined and executed.

An episode is considered successful if the Euclidean distance between the final UAV position $p_{\mathrm{final}}$ and the ground-truth target $p_{\mathrm{gt}}$ is less than a threshold $\delta=20$m:
\begin{equation}
\|p_{\mathrm{final}} - p_{\mathrm{gt}}\|_2 \leq \delta .
\end{equation}

\subsection{Overview of CLOSER-VLN}
As shown in Figure~\ref{fig_2}, CLOSER-VLN comprises three core modules: a hierarchical reasoner, a multidimensional action verifier, and a verification-triggered multimodal retriever. Instead of directly executing the initially predicted action, the framework adopts a pre-execution refinement mechanism. At each time step, the system generates a candidate action and assesses its executability. If verification passes, the action is executed immediately. Otherwise, the system analyzes the error information, generates a targeted query, and retrieves relevant exemplars from the external knowledge base, and refines the action prior to execution.

In our design, the parameters of the hierarchical reasoner are frozen and are not trained to embed relevant knowledge, while the action verifier is trained on data constructed from CityNav\cite{lee2025citynav}, and the knowledge base content is generic and sourced from CityNav. This design differs from conventional retrieval-augmented navigation in two aspects. First, retrieval is not performed unconditionally before every decision but is triggered only by verification failure. Second, retrieved exemplars serve not merely as auxiliary context but as evidence to revise unreasonable actions. Thus, the CLOSER-VLN transforms open-loop action prediction into a bounded closed-loop process of reasoning, verification, retrieval, and refinement.








\subsection{Hierarchical Reasoner}
The hierarchical reasoner serves as the action generation module of CLOSER-VLN. It takes the retrieved supplementary knowledge as input. This design allows CLOSER-VLN to refine decisions within each time step without changing the original task interface.

At time step $t$, let $z_t^k$ denote the supplementary knowledge available at the $k$-th refinement iteration. At the beginning of each time step, no external exemplar is used ($z_t^1=\varnothing$). Specifically, the visual input $v_t$ takes the form of the semantic cognitive map $m_t$ during the navigation and search stages, and switches to the real aerial observation $o_t$ during the localization stage. Given the current stage-level sub-instruction $l_s$, environment description $e_t$, visual input $v_t$, and retrieved knowledge $z_t^k$, the reasoner generates a candidate action $a_t^k$ and its rationale $r_t^k$:
\begin{equation}
(a_t^k, r_t^k) = \mathcal{R}(l_s, e_t, v_t, z_t^k).
\end{equation}

During the navigation and search stages, the reasoner mainly relies on the stage-level sub-instruction, environment description, schematic cognitive map, and retrieved navigation exemplars to output a discrete movement direction. The objective of these stages is to approach the target-related region and construct the target nodes $\mathcal{N}$ according to geospatial and semantic cues.

During the localization stage, the system first predicts target positions via the HSG\cite{xu2026geonav}. If HSG-based retrieval fails (i.e., no eligible nodes $\mathcal{N}$ meet the target criteria), the hierarchical reasoner estimates target coordinates relying on raw aerial observations and retrieved exemplars. This stage aims to achieve fine-grained target localization.

\subsection{Multidimensional Action Verifier}
The multidimensional action verifier is the core module that enables the transition from open-loop action prediction to closed-loop decision optimization.

During the navigation and search phases, we employ a fine-tuned Qwen3-VL-8B \cite{qwen3} as the action verifier, assessing candidate actions across three dimensions. Specifically, instruction consistency checks whether the action aligns with the linguistic command; visual consistency evaluates whether the action is grounded in the semantic cognitive map; and path feasibility determines whether the action leads to a valid path, avoiding backtracking or flying out of map boundaries. The inputs to this module comprise the current stage-level sub-instruction $l_s$, the environment description $e_t$, the semantic cognitive map $m_t$, and the candidate action $a_t^k$ generated by the reasoner. Formally, the verifier outputs a verification rationale $u_t^k$ and a final binary decision of acceptance or rejection $c_t^k$:
\begin{equation}
(u_t^k, c_t^k) = \mathcal{V}(l_s, e_t, m_t, a_t^k),
\end{equation}
where
\begin{equation}
u_t^k = \{u_{\mathrm{ins}}^k, u_{\mathrm{vis}}^k, u_{\mathrm{path}}^k\}
\end{equation}
denotes the verification results for instruction consistency, visual consistency, and path feasibility, respectively.

The final acceptance decision is obtained by aggregating the three verification dimensions:
\begin{equation}
c_t^k = \mathbb{I}[u_{\mathrm{ins}}^k = 1 \land u_{\mathrm{vis}}^k = 1 \land u_{\mathrm{path}}^k = 1],
\label{ck}
\end{equation}
where $c_t^k=1$ indicates that the candidate action is accepted, and $c_t^k=0$ indicates rejection.

During the localization phase, the input to the reasoner differs from the preceding two stages. To mitigate the computational costs associated with further fine-tuning, We conduct a range validity check based on the spatial distance between the predicted coordinate and the current coordinate. Formally, let $\hat{p}$ be the predicted target coordinate, $p_t$ be the current coordinate of the agent, and $\tau_{\text{dist}}$ be the maximum distance threshold (determined by the actual size of the current aerial image, $\tau_{\text{dist}}=50$m \cite{xu2026geonav}). The spatial geometric constraint is formulated as:
\begin{equation}
c_t^k = \mathbb{I} \left( \| \hat{p} - p_t \|_2 \leq \tau_{\text{dist}} \right).    
\end{equation}

In this phase, $c_t^k = 1$ indicates that the predicted target position falls within the valid boundaries of the current aerial observation, allowing the system to confidently accept the localization result. Conversely, if the distance exceeds this threshold ($c_t^k = 0$), implying that the predicted target is hallucinated outside the observable map area, the prediction is deemed unreliable, and an active retrieval process is explicitly triggered to gather further context.

\begin{figure}[t]
\centering
\includegraphics[width=\linewidth]{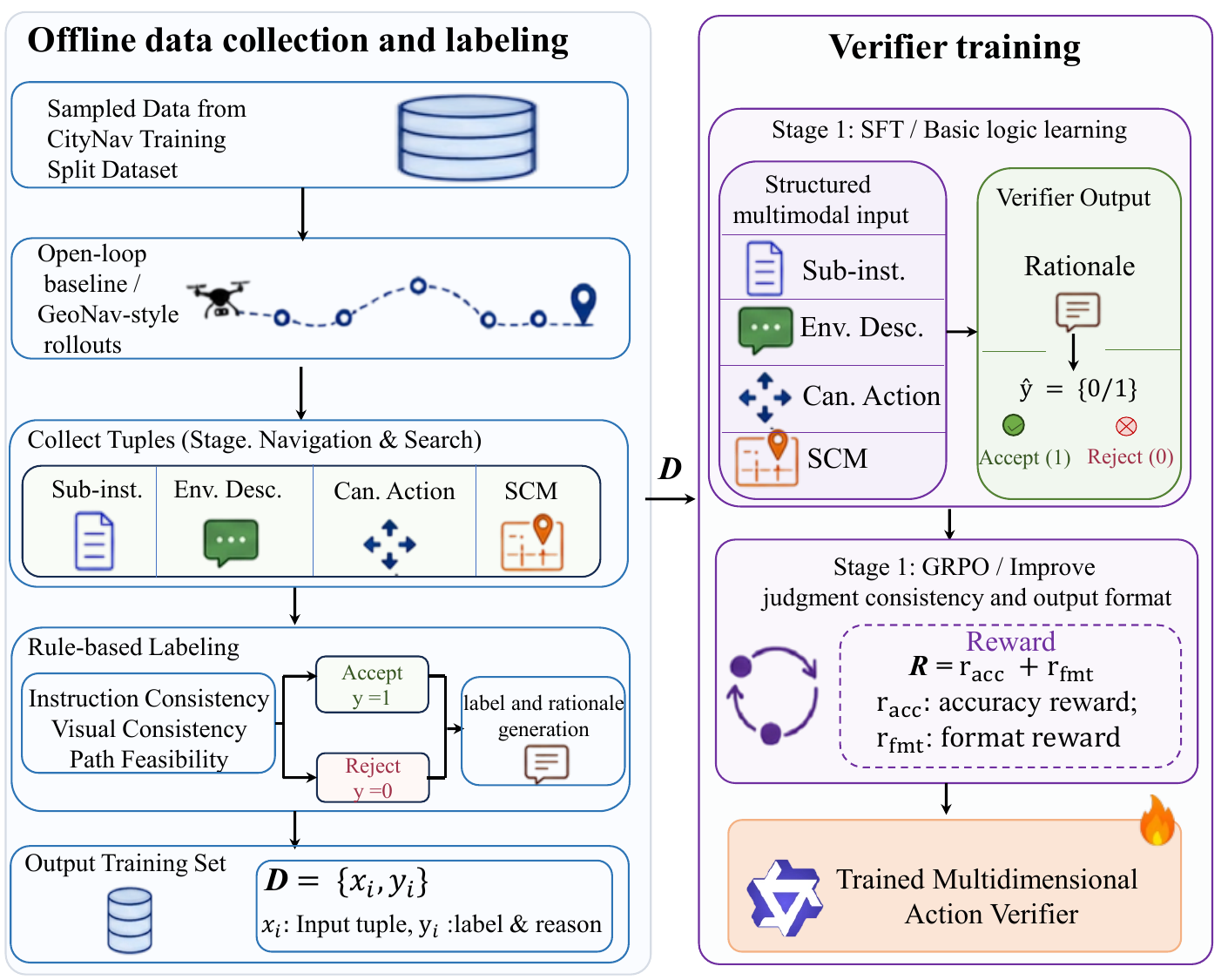}
\caption{Overall pipeline of the multidimensional action verifier. Offline training data are first collected and annotated with labels. A two-stage fine-tuning strategy is then adopted: the model is first trained via SFT to acquire basic logical verification capabilities, followed by reinforcement learning with GRPO to enhance generalization and decision boundary consistency.}
\label{fig_3}
\end{figure}

\subsection{Training of the Action Verifier}
\paragraph{Training Data Construction}
\label{data}
The action verifier is trained as a binary reliability judgment model with structured explanatory outputs. The training data are collected from the CityNav training split~\cite{lee2025citynav}, but they are used only to train the verifier rather than to train a navigation policy. The workflow of data collection and the verifier training is illustrated in
Figure~\ref{fig_3}.

Specifically, we execute the open-loop baseline system on the training split to collect rollout trajectories. For navigation and search steps, we record the sub-instruction $l_{s,i}$, environment description $e_i$, cognitive map $m_i$, and the candidate action $a_i$. We then assign a binary accept/reject label $c_i^*$ based on whether the action meaningfully contributes to the current task stage:
\begin{equation}
c_i^* = \mathbb{I}[u_{\mathrm{ins}}^* = 1 \land u_{\mathrm{vis}}^* = 1 \land u_{\mathrm{path}}^* = 1],
\label{k}
\end{equation}
each label is assigned according to the rule defined in Equation~\ref{k}, ensuring that every correct action points toward the stage goal, relies on the correct visual evidence, and does not result in backtracking or redundant flight paths. Then, to enable the verifier to learn the underlying reasoning logic, we employ Qwen3-VL-32B~\cite{qwen3} to generate a detailed verification rationale $u_i^*$ for each recorded step based on the sample labels. Finally, the dedicated training dataset for the verifier is formulated as follows:

\begin{equation}
\mathcal{D} = \{(x_i, y_i)\}_{i=1}^{N},
\end{equation}
where $x_i = (l_{s,i}, e_i, m_i, a_i)$ represents the structured multimodal input, and $y_i = (u_i^*, c_i^*)$ serves as the ground-truth target comprising the justification and the final verdict.

\paragraph{Fine-Tuning Qwen3-VL-8B}
To improve both the discriminative ability and output stability of the verifier, we adopt a two-stage training strategy.

In the first stage, supervised fine-tuning (SFT) is used to learn the basic logic of action verification. This stage aims to enable the model to fully learn the discriminative rationales within the dataset, making it an effective verification model. To apply the standard causal language modeling objective, the structured target $y_i = (u_i^*, c_i^*)$ is first serialized into a natural language sequence of discrete tokens, denoted as $Y_i = (y_{i,1}, y_{i,2}, \dots, y_{i, L})$. The supervised fine-tuning objective is to minimize the negative log-likelihood:
\begin{equation}
\mathcal{L}_{\mathrm{SFT}}(\theta)
=
-
\sum_{(x_i,y_i)\in\mathcal{D}}
\sum_{j=1}^{|Y_i|}
\log P_{\theta}(y_{i,j}|x_i,Y_{i,<j}).
\end{equation}

In the second stage, we adopt Group Relative Policy Optimization (GRPO) to further improve judgment consistency and structured output quality\cite{grpo-start}. For each input prompt, the model samples a group of candidate outputs and is optimized according to relative rewards. Our dataset, as well as the inference process of the navigation task, suffers from inherent class imbalance, where positive accept labels account for only about 40\%. To avoid reward hacking, we design an expectation-based asymmetric reward function. It constrains the model to strictly reject invalid actions and enables the model to effectively learn the multimodal verification task. The final reward consists of a soft format penalty and an asymmetric accuracy reward:
\begin{equation}
R = r_{\mathrm{acc}} + r_{\mathrm{fmt}}.
\end{equation}

The format reward $r_{\mathrm{fmt}}$ acts as a soft penalty to encourage machine-parseable structured outputs. Rather than terminating the reward calculation entirely upon a format violation, we apply a moderate penalty to maintain stable gradient signals:
\[
r_{\mathrm{fmt}} =
\begin{cases}
1.0, & \text{the output follows the required format}, \\
-1.5, & \text{otherwise}.
\end{cases}
\]

The accuracy reward $r_{\mathrm{acc}}$ assigns fine-grained values based on the four outcomes of the confusion matrix. False Positives (FP) are heavily penalized to strictly prevent the system from accepting hallucinated or incorrect reasoning. False Negatives (FN) receive a slightly lower penalty to deter the model from adopting a zero-risk all-reject policy. True outcomes receive high positive rewards to incentivize genuine classification:
\[
r_{\mathrm{acc}} =
\begin{cases}
2.0, & \hat{y}=0 \text{ and } y^*=0 \text{ (True Negative)}, \\
1.5, & \hat{y}=1 \text{ and } y^*=1 \text{ (True Positive)}, \\
-3.0, & \hat{y}=0 \text{ and } y^*=1 \text{ (False Negative)}, \\
-4.0, & \hat{y}=1 \text{ and } y^*=0 \text{ (False Positive)},
\end{cases}
\]
where $\hat{y} \in \{0, 1\}$ is the predicted judgment parsed from the model output (corresponding to the online verification decision $c_t^k$), and $y^* \in \{0, 1\}$ is the ground-truth label (corresponding to $c_i^*$ in the training dataset). This reward encourages the model to produce a valid verification rationale and a final accept/reject decision.

Through the meticulous two-stage design described above\cite{sft-grpo}, the resulting Qwen3-VL-8B verifier not only demonstrates excellent discriminative performance on offline test sets, but more importantly, its structured output decisions can be seamlessly parsed and executed by the downstream closed-loop pipeline. The first stage ensures that the model fully comprehends the task semantics, while the second stage employs an asymmetric reward mechanism to enforce a conservative rejection tendency when the model is uncertain, thereby substantially reducing risk behaviors caused by hallucinations in practical navigation tasks. This training paradigm offers a viable pathway for deploying vision-language models in safety-critical robotic applications.

\begin{figure*}[t]
\centering
\includegraphics[width=\linewidth]{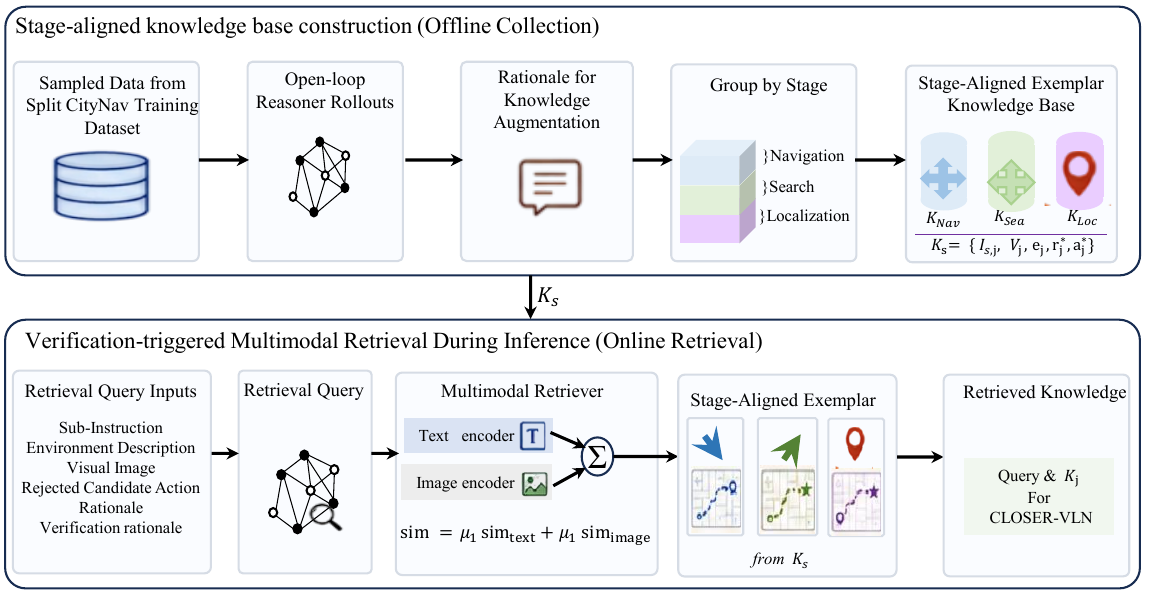}
\caption{Verification-triggered multimodal retrieval. The exemplar knowledge base is constructed from successful decision fragments and organized by task stage and spatial semantics. During inference, retrieval is activated only after verification failure, and the retrieved exemplar provides corrective evidence for action refinement.}
\label{fig_4}
\end{figure*}

\subsection{Verification-Triggered Multimodal Retriever}
Unlike conventional methods that retrieve auxiliary knowledge before every inference step, the CLOSER-VLN triggers retrieval only when an action is rejected: the verification-triggered multimodal retriever generates a query from the erroneous information and retrieves relevant exemplars from an external knowledge base to supplement task knowledge. Thus, retrieval in the CLOSER-VLN is not a general context enhancement mechanism but an error-aware and on-demand correction mechanism. The construction and supplementation of the external knowledge base are shown in Figure~\ref{fig_4}.

To build a representative external knowledge base, we run the open-loop policy on the training dataset and sample episodes from the resulting trajectories, and categorize them according to the number of landmarks and the types of target objects. Each unique combination of landmark count and object category is treated as an independent case type, and at least five trajectories are stored for each type in the database. For each optimal decision rationale $r_j^*$, it is also required to specify the current position $(x, y)$, the target bounding box $[X_{\min}, Y_{\min}, X_{\max}, Y_{\max}]$, and the justification for selecting that target, thereby ensuring the correctness and sufficiency of the knowledge in the retrieved exemplars. Finally, the retained cases are grouped according to task stage, target category, number of landmarks, and spatial-relation type, and are stored in the navigation, search, and localization subsets.

Based on the database constructed above, we define a stage-aligned exemplar knowledge base:
\begin{equation}
\mathcal{K}_s, \quad s \in \{\mathrm{Nav}, \mathrm{Sea}, \mathrm{Loc}\}.
\end{equation}

To clearly distinguish the verified training data from the training dataset, we use the subscript $j$ for knowledge base indexing and the superscript $*$ to denote optimal or corrected labels. Each exemplar is represented as a structured tuple:
\begin{equation}
\mathcal{K}_{s,j} = \langle e_j, v_j, l_{s,j}, r_j^*, a_j^* \rangle,
\end{equation}
where $e_j$ denotes the environment description, $v_j$ denotes the visual image ($m_j$ in the $\mathcal{K}_{Nav}$ and $\mathcal{K}_{Sea}$, $o_j$ in the $\mathcal{K}_{Loc}$) representation, $l_{s,j}$ denotes the sub-instruction, $r_j^*$ denotes the optimal decision rationale, and $a_j^*$ denotes the verified action or action sequence.

When the $k$-th candidate action is rejected by the verifier, the retrieval analyzer generates a query according to the current context and the failed verification dimensions:
\begin{equation}
q_t^k = \mathcal{Q}(l_s, e_t, v_t, a_t^k, r_t^k, u_t^k).
\end{equation}

The retrieval query is constructed by integrating the stage-level sub-instruction, the count and categories of task-relevant landmarks, the specific knowledge demands inferred from the verifier's rationale, the target object category, and the suspected failure type associated with the rejected action.

The retrieval process first identifies the optimal exemplar $E^*$ by maximizing the multimodal similarity:
\begin{equation}
E^* = \arg\max_{E_j \in \mathcal{K}_s} \operatorname{sim} \left( \phi_q(q_t^k, v_t), \phi_E(E_j) \right),
\end{equation}
where $\phi_q$ and $\phi_E$ denote the multimodal representations of the query and exemplar, respectively. The search space is restricted to $\mathcal{K}_s$ with $s$ fixed to the current task stage, i.e., $\mathcal{K}_{\mathrm{nav}}$ during navigation, $\mathcal{K}_{\mathrm{sea}}$ during search, and $\mathcal{K}_{\mathrm{loc}}$ during localization. Cross-stage retrieval is disallowed. The similarity function is defined as a weighted combination of textual and map-structural similarities:
\begin{equation}
\operatorname{sim} = \mu_1 \cdot \operatorname{sim}_{\mathrm{text}} +  \mu_2 \cdot \operatorname{sim}_{\mathrm{image}}.
\end{equation}

To balance the consideration of semantic meaning and spatial topology during the retrieval process, we set $\mu_1 = \mu_2 = 0.5$. The textual similarity $\operatorname{sim}_{\mathrm{text}}$ matches instruction and the suspected failure type from the query context. The visual observation similarity $\operatorname{sim}_{\mathrm{image}}$ matches environmental layouts, landmark appearances, and spatial configurations.

Ultimately, the retrieved knowledge $z_t^{k+1}$ is formally constructed by concatenating the retrieved optimal exemplar and the error-aware query:
\begin{equation}
z_t^{k+1} = (E^*, q_t^k).
\end{equation}

This combined knowledge $z_t^{k+1}$ is then transformed into a structured prompt and fed into the hierarchical reasoner for the next refinement iteration.

\subsection{Bounded Closed-Loop Inference}

\begin{algorithm}[t]
\caption{Bounded closed-loop inference of CLOSER-VLN}
\label{alg:CLOSER_inference}
\begin{algorithmic}[1]
\REQUIRE sub-instruction $l_s$, environment description $e_t$, observation $o_t$, cognitive map $m_t$, stage-aligned knowledge base $\mathcal{K}_s$, maximum iteration number $K_{\max}$
\ENSURE Executed action $a_t$
\STATE Initialize retrieved knowledge $z_t^1 \leftarrow \varnothing$
\FOR{$k = 1$ to $K_{\max}$}
    \STATE Generate candidate action and rationale using the hierarchical reasoner
    \STATE Evaluate the candidate action using the multidimensional verifier
    \IF{the candidate action is accepted}
        \STATE Execute the candidate action
        \STATE \textbf{return} executed action
    \ELSE
        \STATE Generate an error-aware retrieval query
        \STATE Retrieve a stage-aligned exemplar from $\mathcal{K}_s$
        \STATE Feed the retrieved exemplar back to the reasoner
    \ENDIF
\ENDFOR
\STATE Execute the final candidate action as fallback
\STATE \textbf{return} executed action
\end{algorithmic}
\end{algorithm}

During inference, the CLOSER-VLN performs bounded closed-loop refinement before environment step execution. At each time step, the retrieved knowledge is initialized as empty. The reasoner first generates a candidate action and its rationale. The verifier then evaluates the reliability of the candidate action. If the candidate is accepted, it is immediately executed. If it is rejected, the retrieval analyzer generates an error-aware query, retrieves a stage-aligned exemplar from the external knowledge base, and feeds both retrieved knowledge and error analysis back to the reasoner for the next refinement iteration, as shown in Algorithm~\ref{alg:CLOSER_inference}.

This process is repeated until a candidate action is accepted or the maximum number of refinement iterations is reached. In our implementation, the maximum number of refinement iterations is set to $K_{\max}=3$. If no candidate action is accepted within this limit, the CLOSER-VLN executes the final candidate action generated in the last iteration as a fallback. This strategy ensures that the closed-loop refinement process improves decision reliability while keeping inference cost bounded.

\section{Experiments}




\begin{table*}[t]
    \centering
    \begin{threeparttable}
        \caption{Performance comparison on Test Unseen, Validation Unseen, and Validation Seen}
        \label{main_result}

        \small 
        
        \setlength{\tabcolsep}{6pt} 

        \begin{tabular}{l cccc cccc cccc}
            \toprule
            \multirow{2}{*}{\textbf{Method}} & \multicolumn{4}{c}{\textbf{Validation Seen}} & \multicolumn{4}{c}{\textbf{Validation Unseen}} & \multicolumn{4}{c}{\textbf{Test Unseen}} \\
            \cmidrule(lr){2-5} \cmidrule(lr){6-9} \cmidrule(lr){10-13}
            & NE$\downarrow$ & SR$\uparrow$ & OSR$\uparrow$ & SPL$\uparrow$ & NE$\downarrow$ & SR$\uparrow$ & OSR$\uparrow$ & SPL$\uparrow$ & NE$\downarrow$ & SR$\uparrow$ & OSR$\uparrow$ & SPL$\uparrow$ \\
            \midrule
            Random           & 222.30 & 0.00 & 0.00 & 0.00 & 223.00 & 0.00 & 0.90 & 0.00 & 208.80 & 0.00 & 1.44 & 0.00 \\
            Seq2Seq         & 148.40 & 4.52 & 10.61 & 4.47 & 201.40 & 1.04 & 8.03 & 1.02 & 174.50 & 1.73 & 8.57 & 1.69 \\
            CMA             & 151.70 & 3.74 & 10.77 & 3.70 & 205.20 & 1.08 & 7.89 & 1.06 & 179.10 & 1.61 & 10.07 & 1.57 \\
            MGP               & 59.70 & 8.69 & 10.55 & 35.51 & 75.10 & 5.84 & 22.19 & 5.56 & 93.80 & 6.38 & 26.04 & 6.08 \\
            FlightGPT \cite{cai2025flightgpt}       & 66.10 & 17.57 & 30.26 & 15.78 & 68.10 & 14.69 & 29.33 & 13.24 & 76.20 & 21.20 & 35.38 & 19.24 \\
            \midrule
            GeoNav \cite{xu2026geonav}     & \textbf{59.41}  & 22.04 & 38.37 & 12.87 & \textbf{63.11} & 16.41 & 35.55 & 9.85 & 73.69 & 25.90 & 41.66 & 16.03 \\
            \textbf{CLOSER-VLN (Ours)} & 65.30 & \textbf{30.78} & \textbf{40.90} & \textbf{19.54} & 66.57 & \textbf{21.44} & \textbf{37.83} & \textbf{14.64} & \textbf{71.42} & \textbf{32.01} & \textbf{43.51} & \textbf{21.28} \\
            \bottomrule
        \end{tabular}
        
    \end{threeparttable}
\end{table*}

\begin{table*}[t]
    \centering
    \begin{threeparttable}
        \caption{Performance comparison on Validation seen. \textbf{bold} indicates best.}
        \label{results}

        \setlength{\tabcolsep}{6pt} 
        \small 

        \begin{tabular}{l cccc cccc cccc}
            \toprule
            \multirow{2}{*}{\textbf{Method}} & \multicolumn{4}{c}{\textbf{Easy}} & \multicolumn{4}{c}{\textbf{Medium}} & \multicolumn{4}{c}{\textbf{Hard}} \\
            \cmidrule(lr){2-5} \cmidrule(lr){6-9} \cmidrule(lr){10-13}
            & NE$\downarrow$ & SR$\uparrow$ & OSR$\uparrow$ & SPL$\uparrow$ & NE$\downarrow$ & SR$\uparrow$ & OSR$\uparrow$ & SPL$\uparrow$ & NE$\downarrow$ & SR$\uparrow$ & OSR$\uparrow$ & SPL$\uparrow$ \\
            \midrule
            Random  & 340.62 & 0.00 & 0.00 & 0.00 & 548.30 & 0.00 & 0.00 & 0.00 & 654.26 & 0.00 & 0.00 & 0.00 \\
            Greedy  & 105.63 & 1.55 & 3.15 & 0.97 & 103.00 & 0.00 & 4.88 & 0.00 & 78.80 & 0.00 & 2.71 & 0.00 \\
            GPT-4o  & 301.45 & 0.00 & 7.50 & 0.00 & 327.25 & 0.00 & 6.25 & 0.00 & 401.63 & 0.00 & 0.00 & 0.00 \\
            \midrule
            GeoNav \cite{xu2026geonav} & 59.86 & 26.53 & \textbf{73.47} & 12.05 & \textbf{53.80} & 22.92 & 39.58 & 17.06 & \textbf{68.90} & 16.67 & 22.92 & 12.49 \\

            \textbf{CLOSER-VLN (Ours)} & \textbf{53.89} & \textbf{40.67} & 52.87 & \textbf{25.68} & 62.01 & \textbf{29.74} & \textbf{41.02} & \textbf{19.15} & 79.57 & \textbf{22.11} & \textbf{29.12} & \textbf{13.90} \\
            \bottomrule
        \end{tabular}

    \end{threeparttable}
\end{table*}

\subsection{Experimental Setup}

\subsubsection{Dataset and Splits}

The proposed framework is evaluated on the CityNav benchmark dataset, which provides bird's-eye view images and prior landmark information to effectively measure UAV performance in landmark-based visual target retrieval. The dataset is divided into three standard splits: seen validation, unseen validation, and unseen test sets. To comprehensively evaluate model capability, navigation tasks are categorized into three difficulty levels based on the Euclidean distance from start to target, enabling analysis of flight distance impact on aerial navigation performance.
\subsubsection{Evaluation Metrics}
To quantitatively measure navigation performance from multiple perspectives, we employ four primary evaluation metrics:

\begin{itemize}
    \item Navigation Error (NE): Measures the average Euclidean distance between the agent's final stopping position and the ground-truth target.
    \item Success Rate (SR): The percentage of tasks in  the agent successfully completes navigation within a 20-meter Euclidean distance of the ground-truth target.
    \item Oracle Success Rate (OSR): Evaluates whether the agent's position was within during the entire trajectory.
    \item Success weighted by Path Length (SPL): A measure of efficiency that weights the success rate by the ratio of the reference path length to the actual path length.
\end{itemize}

\subsubsection{Implementation Details}

We employ GPT-4o as the hierarchical reasoner \cite{xu2026geonav} and also use it to implement the retrieval analyzer (i.e., generating error-aware queries), while Qwen3-VL-8B serves as the verifier. We set the sampling temperature to 0 for all MLLMs. The training configuration of the verifier is summarized in Table~\ref{tab:training_configs}. Experiments are conducted on a server equipped with two Intel Xeon Platinum 8375C CPUs and four NVIDIA A100 (80GB) GPUs. The software environment includes vLLM v0.11.0, DeepSpeed ZeRO-3, and bfloat16.Baseline Models

\subsubsection{Baseline Models}
We compare the proposed CLOSER-VLN with representative methods on the CityNav benchmark, including supervised baselines (Seq2Seq, CMA, MGP, FlightGPT) and zero-shot baselines (GeoNav, Greedy, GPT-4o). To better delineate our contributions, We summarize in Table~\ref{tab:information} summarizes the reasoning paradigm (open-loop or closed-loop) adopted by each baseline method, along with their key technical configurations, including the application of pretrained models, SCM, HSG, and knowledge retrieval (BS), as well as the information used to construct our verifier. Among them, Seq2Seq, CMA, MGP, and FlightGPT are trained on the CityNav dataset \cite{lee2025citynav}, whereas our verifier is trained on the dataset constructed in Section~\ref{data}.

\begin{table}[t]
    \centering
    \caption{Information usage of different methods, BS denotes knowledge retrieval}
    \label{tab:information}
    \small 
    \setlength{\tabcolsep}{4pt} 
    \renewcommand{\arraystretch}{1.2}
    \begin{tabular}{lccccc}
        \toprule
        \textbf{Method} & \textbf{Closed-loop} & \textbf{Pre-train} & \textbf{SCM} & \textbf{HSG} & \textbf{BS} \\ 
        \midrule
        Seq2Seq    & \xmark & \cmark & \xmark & \xmark & \xmark \\
        CMA        & \xmark & \cmark & \xmark & \xmark & \xmark \\
        MGP        & \xmark & \cmark & \xmark & \xmark & \xmark \\
        FlightGPT  & \xmark & \cmark & \xmark & \xmark & \xmark \\
        Greedy     & \xmark & \xmark & \xmark & \xmark & \xmark \\
        GPT-4o     & \xmark & \xmark & \xmark & \xmark & \xmark \\
        GeoNav     & \xmark & \xmark & \cmark & \cmark & \xmark \\
        CLOSER-VLN & \cmark & \xmark & \cmark & \cmark & \cmark \\
        Verifier   & \xmark & \cmark & \cmark & \xmark & \xmark \\
        \bottomrule
    \end{tabular}
\end{table}

\begin{table}[t]

    \centering

    \caption{TRAINING HYPER-PARAMETERS FOR SFT AND GRPO}

    \label{tab:training_configs}

    \setlength{\tabcolsep}{14pt} 

    \renewcommand{\arraystretch}{1.2}

    \begin{tabular}{lccc}

        \toprule

        \textbf{Stage} & \textbf{Batch Size} & \textbf{LR} & \textbf{Epochs} \\ 

        \midrule

        SFT  & 4 & $1.0 \times 10^{-5}$ & 3 \\

        GRPO & 1 & $1.0 \times 10^{-6}$ & 1 \\

        \bottomrule

    \end{tabular}

\end{table}

\subsection{Main Results}
Table \ref{main_result} reports the results on the standard Validation Seen, Validation Unseen, and Test Unseen splits, while Table \ref{results} presents the performance under different navigation difficulty levels. Overall, CLOSER-VLN achieves the best success-related performance among all methods, demonstrating the effectiveness and strong generalization capability of the proposed closed-loop spatial reasoning framework.

\paragraph{Performance Across Seen and Unseen Splits}

As shown in Table~\ref{main_result}, supervised aerial VLN methods deliver inferior overall performance on the Test Unseen split. Unlike ground navigation, aerial navigation features long travel distances and drastic view changes, which create greater reliance on long-term decision-making. A single deviation in intermediate decisions will cause continuous error accumulation, eventually leading to failure in target localization. Although FlightGPT and GeoNav achieve decent performance by leveraging the powerful vision-language understanding capabilities of large models, their performance is limited by the strategy of executing decisions after one-time generation, and they cannot autonomously detect and correct errors in intermediate steps during reasoning. In contrast, the proposed CLOSER-VLN achieves relative improvements of 23.6\% in SR and 32.8\% in SPL over the best-performing zero-shot baseline, GeoNav. This result indicates that CLOSER-VLN not only attains a higher navigation success rate but also generates more efficient trajectories, which fully demonstrates the advantages of the iterative reasoning and self-correction mechanism over the one-pass decision generation paradigm.

A similar trend is observed on the Validation Unseen split. Compared with GeoNav, the CLOSER-VLN increases SR from 16.41\% to 21.44\% and SPL from 9.85\% to 14.64\%. On the Validation Seen split, the CLOSER-VLN reaches 30.78\% in SR, 40.90\% in OSR and 19.54\% in SPL. This demonstrates that the proposed closed-loop reasoning paradigm enables effective decision error correction across different environments.

Notably, the GeoNav obtains slightly lower NE on both Validation Seen and Validation Unseen splits, yet this advantage does not translate into a higher navigation success rate. It reveals that affected by unrecoverable reasoning errors in its open-loop decision process, GeoNav often terminates navigation near the target without meeting the stopping criteria. By comparison, the slight increase in NE of the CLOSER-VLN is attributed to the fact that the agent conducts further exploration and keeps searching for the target in failed cases, instead of terminating the task prematurely.

\paragraph{Performance Across Difficulty Levels}

Table~\ref{results} further presents the performance of all methods categorized by difficulty on the Validation Seen split. CLOSER-VLN achieves the highest success rate across all difficulty levels. Specifically, it attains SR of 40.67\%, 29.74\%, and 22.11\% on easy, moderate, and hard tasks, respectively, corresponding to relative improvements of 53.3\%, 29.8\%, and 32.6\% over GeoNav, with a particularly notable gain for hard tasks. This is because hard tasks impose strict requirements on long-distance planning and precise landmark grounding. Open-loop methods fail to revise erroneous decisions once made and are prone to error accumulation. In contrast, CLOSER-VLN continuously verifies decision reliability, corrects faulty predictions, and sustains stable decision-making capability. Moreover, CLOSER-VLN yields improved success weighted by path length at all difficulty levels, indicating that it boosts the success rate while ensuring path efficiency. The above results verify that the proposed closed-loop reasoning strategy delivers favorable performance gains across different difficulty levels.

Since we employ VLMs as the core reasoner in this work, we conduct multiple independent experiments on the easy difficulty level of the seen split in CLOSER-VLN to validate the stability of our method against the inherent stochasticity of VLMs. Specifically, the standard deviation of SR is merely 0.6\%, and the standard deviation of SPL is 0.8\%. Such minimal variance verifies that the closed-loop reasoning mechanism proposed in this paper effectively restricts the reasoning pipeline of vision-language models, greatly alleviates the lucky-seed performance bias induced by VLMs, and ensures highly reproducible navigation experimental results.

\subsection{Component Ablation}
Table~\ref{tab:ablation} reports the ablation results of each core component of the CLOSER-VLN on the unseen test set. Taking the variant with only the reasoning module (Reasoner only) as the baseline, it achieves a SR of 24.76\% and a SPL of 14.87\%, with an NE of 85.03m. After introducing the verifier module (+Verifier), the SR increases to 29.24\%, the SPL to 19.09\%, and the NE decreases to 77.26, indicating that explicit intermediate decision verification can effectively mitigate the error accumulation problem in open-loop reasoning. However, this variant remains limited by the insufficient knowledge of the reasoning module itself, making it difficult to make optimal decisions within a limited number of iterations. In contrast, introducing simple retrieval at every step (+Always Simple Retrieval) yields only marginal gains (SR of 26.96\%) and even causes NE to deteriorate to 97.25 due to the introduction of irrelevant information in some retrieval steps. Combining verification with simple retrieval (+Verifier + Simple Retrieval) further improves SR to 30.28\% and SPL to 19.89\%. However, due to the lack of goal-oriented retrieval, a small amount of task-irrelevant knowledge is introduced, resulting in lower performance than the full model.. Finally, the complete CLOSER-VLN (with the verifier and adaptive target knowledge retrieval) achieves the best results: SR of 32.01\%, SPL of 21.28\%, and NE of 71.42. These results collectively demonstrate that the verifier module and the target knowledge retrieval module have a synergistic effect: the verifier is responsible for validating the reliability of the reasoning output, while the retrieval is responsible for supplying critical spatial knowledge.

\begin{table}[t] 
\centering
\footnotesize  
\caption{The ablation study on Test Unseen of CityNav}
\label{tab:ablation}
\resizebox{\columnwidth}{!}{
    \setlength{\tabcolsep}{1.2em}
    \begin{tabular}{lcccc}
    \toprule
    \textbf{Config}  & \textbf{NE}$\downarrow$ & \textbf{SR}$\uparrow$  & \textbf{OSR}$\uparrow$  & \textbf{SPL}$\uparrow$  \\
    \midrule
Reasoner only       & 85.03 & 24.76 & 33.35 & 14.87 \\
+ Verifier       &  77.26 & 29.24 & 39.31 & 19.09     \\
+ Always Simple Retrieval & 97.25  & 26.96 & 36.57& 16.53 \\
+ Verifier + Simple Retrieval & 75.59& 30.28 &41.37 & 19.89    \\
\textbf{Full CLOSER-VLN}    &  \textbf{71.42}  & \textbf{32.01}  &\textbf{43.51}
& \textbf{21.28}    \\
\bottomrule
\end{tabular}
}
\end{table}

\subsection{Verifier Analysis}
To verify the necessity of fine-tuning a dedicated verifier, we evaluate its judgment capability on an independently constructed evaluation set, which is derived from the Validation Seen split of the CityNav dataset (see Section \ref{data} for the construction methodology). This evaluation set contains 387 positive samples and 613 negative samples. In Aerial VLN tasks, failing to intercept erroneous actions (i.e., miss-rejection) often causes the agent to deviate from its trajectory or even leads to irreversible task failure. Therefore, our analysis focuses on the verifier's ability to "Reject" actions that are actually "Wrong," primarily examining the rejection precision and rejection recall.

\begin{figure}[t]
\centering
\includegraphics[width=\linewidth]{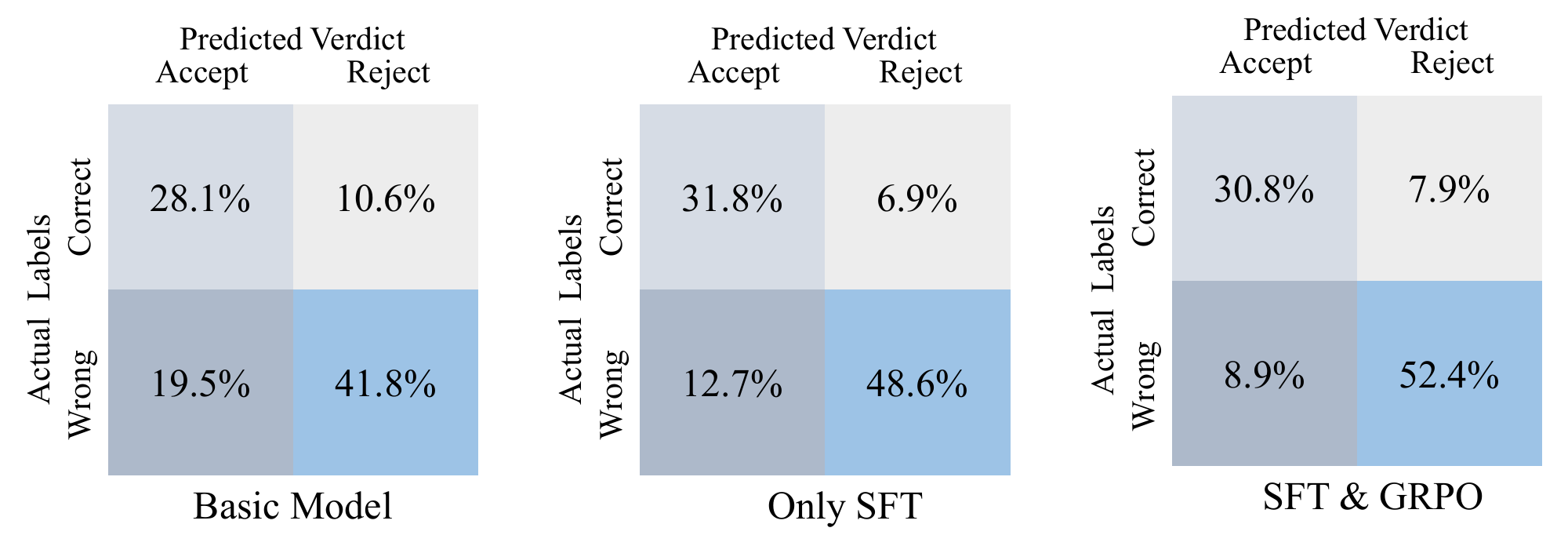}
\caption{Confusion matrices of different verifiers. Rows represent ground-truth labels (whether the action is actually Correct or Wrong), and columns represent predicted verdicts (whether the verifier judges the action as Accept or Reject). Each cell reports the proportion of all evaluation samples falling into that category. The three subfigures correspond to the basic model, the SFT-only model, and the full SFT+GRPO model, respectively.}
\label{fig_5}
\end{figure}

The confusion matrices of different verifiers are shown in Figure~\ref{fig_5}. From the rejection perspective, the following performance evolution can be observed:

Base Model exhibits a severe "over-trust" tendency. When facing wrong actions, its rejection recall is only 68.2\% (41.8\% / 61.3\%), meaning that nearly one-third of erroneous actions are blindly accepted by the model (with a miss-rejection rate as high as 19.5\%). Meanwhile, its rejection precision is 79.8\% (41.8\% / 52.4\%). The overall judgment capability struggles to meet the reliability requirements of complex navigation tasks.

SFT Model injects basic judgment norms into the model, but the decision boundary remains coarse. After SFT, the model is infused with specialized judgment knowledge, significantly improving its rejection precision to 87.6\% (48.6\% /55.5\%) and boosting recall to 79.3\% (48.6\% / 61.3\%). However, relying solely on SFT still has limitations. When handling ambiguous or borderline samples, the model faces a trade-off dilemma, and the 12.7\% miss-rejection rate indicates that there is still room for improvement in its ability to intercept potential risks.

The SFT+GRPO model further optimizes the decision boundary and significantly enhances verification performance. After introducing GRPO fine-tuning, the model achieves improved performance in rejection capability, with a rejection recall of 85.5\% (52.4\% / 61.3\%) and a substantially reduced miss-rejection rate of 8.9\%, greatly lowering the probability of fatal errors. Although the reward function design bias leads to a slight sacrifice in precision (marginally decreasing to 86.9\%) when strictly intercepting errors, the substantial gain in recall achieved by the complete two-stage fine-tuned model ensures the credibility of closed-loop inference during navigation.

\subsection{Efficiency Analysis}

We further investigate the impact of the maximum number of iterations per time step, denoted as $K$, on system performance. Here, $K$ represents the maximum number of decisions that the reasoner can generate within one time step. When $K_{\max}=1$, the system degenerates to an open-loop strategy. As shown in Table~\ref{tab:blation}, as $K$ increases from 1 to 3, the model performance improves continuously, reaching the optimum at $K_{\max}=3$ with a SR of 32.01\%, a SPL of 21.28\%, and a NE of 71.42. However, when $K$ is further increased to 4, the SR exhibits a clear decline. The reason for this phenomenon is that, after multiple iterations, the erroneous analysis $q_t^k$ produced by the retrieval module introduces additional analytical information, some of which is irrelevant or even contradictory. This extra noise interferes with the decision-making process of the reasoning module, not only incurring additional system overhead but also causing its negative impact to outweigh the benefits gained from multiple iterations. Therefore, we set $K_{\max}=3$ as the default configuration for the CLOSER-VLN.

\begin{table}[t] 
\centering
\caption{The impact of maximum refinement iterations on Test Unseen of CityNav}
\label{tab:blation}
\resizebox{\columnwidth}{!}{
    \setlength{\tabcolsep}{1.2em} 
    \begin{tabular}{lcccc}
    \toprule
    \textbf{No. of refinement iterations}& \textbf{NE}$\downarrow$ & \textbf{SR}$\uparrow$ & \textbf{OSR}$\uparrow$ &\textbf{SPL}$\uparrow$  \\
    \midrule
K = 1    & 85.03 & 24.76 &33.35 & 14.87 \\
K = 2    &80.93 & 28.64  &39.56 & 18.04  \\
K = 3      & \textbf{71.42} & \textbf{32.01}&43.51  & \textbf{21.28} \\
K = 4    & 76.46 & 30.18   & \textbf{46.86} &20.43\\

\bottomrule
\end{tabular}
}
\end{table}

\begin{figure*}[t]
\centering
\includegraphics[width=\linewidth]{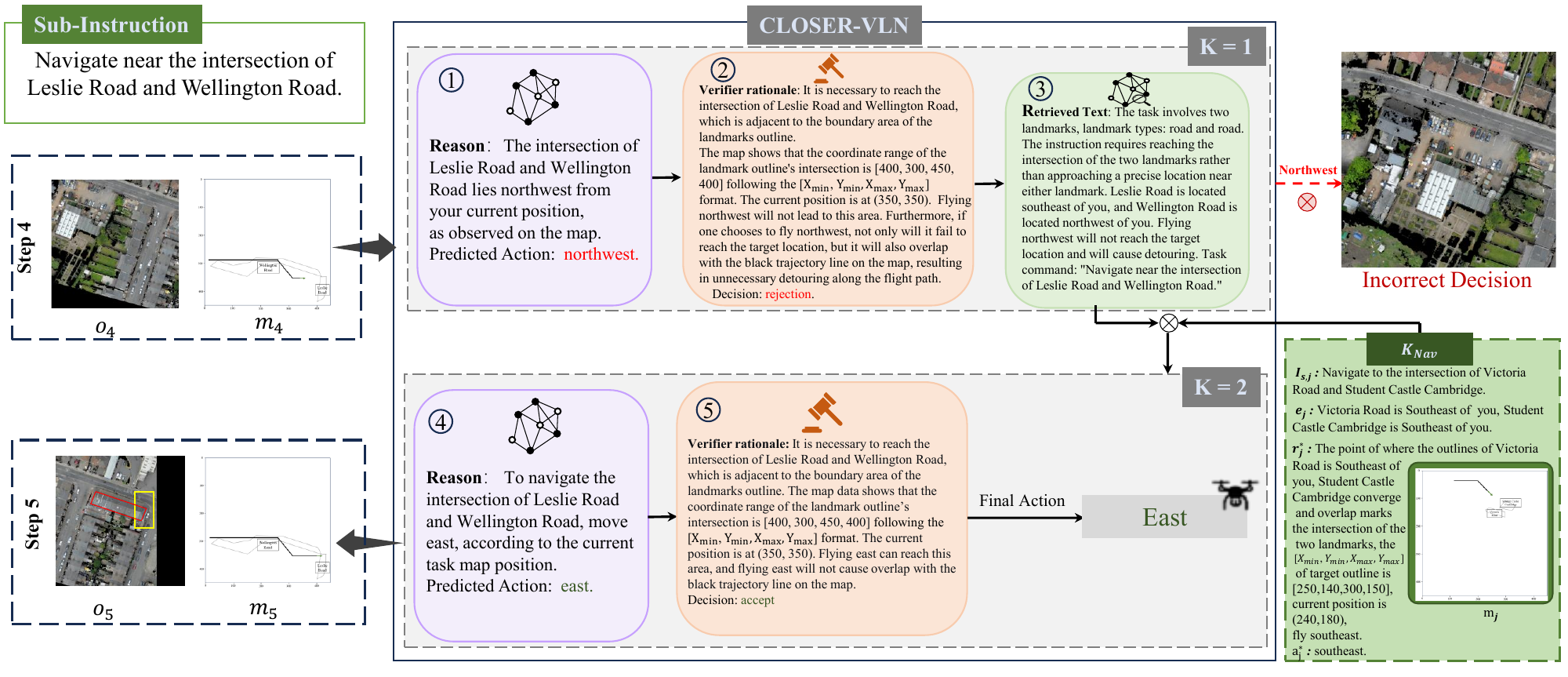}
\caption{A visualization of the system's explicit closed-loop reasoning during a UAV navigation episode in the CityNav environment. The CLOSER-VLN takes the SCM  ($m_4$) and the subtask instruction as inputs. At $k=1$, the multidimensional action verifier rejects the incorrect action (northwest) proposed by the hierarchical reasoner, deducing that it fails to advance the task and causes detouring. Subsequently, the verification-triggered multimodal retriever extracts a successful historical case. This retrieved knowledge is incorporated into the next reasoning iteration ($k=2$), leading to the correct action (east). Consequently, in Step 5 ($o_5$), the UAV successfully approaches both the stage goal (yellow box) and the final goal (red box).}
\label{fig_6}
\end{figure*}

\begin{figure*}[t]
\centering
\includegraphics[width=\linewidth]{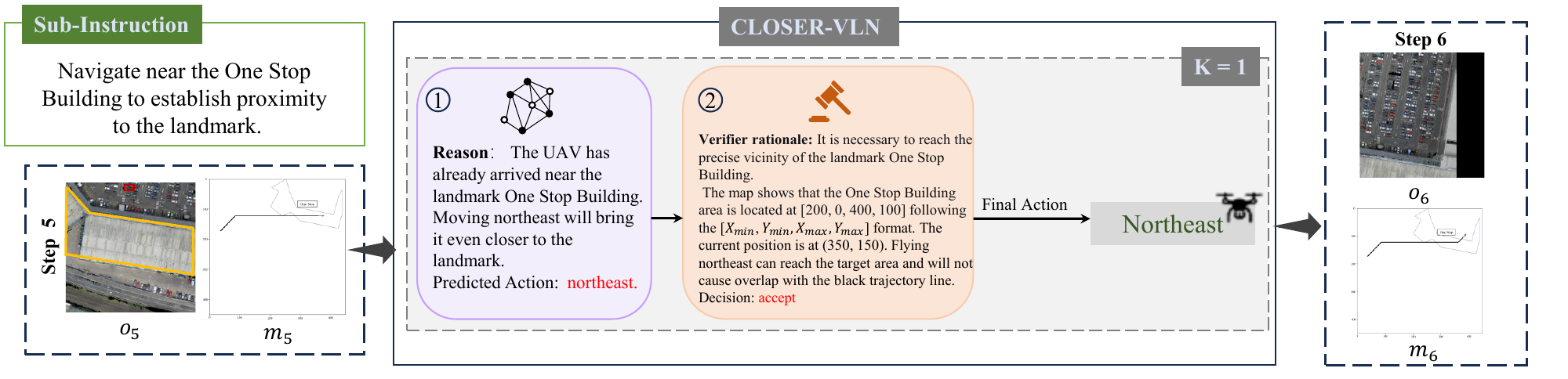}
\caption{A visualization of a verification failure case during a UAV navigation episode in the CityNav environment. The CLOSER-VLN takes the SCM ($m_4$) and the subtask instruction as inputs. At $k=1$, the multidimensional action verifier erroneously accepts the incorrect action proposed by the hierarchical reasoner, failing to recognize the complex geometric contours and peripheral boundary of the target landmark (the One Stop Building). This verification failure causes the UAV to bypass the optimal subtask termination zone without halting. Consequently, while the UAV successfully traverses the destination area in Step~5 ($o_5$), it fails to stop in Step~6 ($o_6$).}
\label{fig_7}
\end{figure*}

\subsection{Qualitative Analysis}
\subsubsection{Successful Case: Closed-Loop Correction}
To demonstrate the efficacy of the CLOSER-VLN closed-loop reasoning paradigm in Unmanned Aerial Vehicle Visual-and-Language Navigation (UAV-VLN), we present a step-level analysis of a successful reasoning episode within the CityNav environment (see Figure~\ref{fig_6}). Given the sub-instruction ``Navigate near the intersection of Leslie Road and Wellington Road,'' the system must make a directional decision at Step 4. 

During the initial reasoning iteration ($k=1$), the hierarchical reasoner generates a preliminary decision to fly ``northwest.'' However, due to an incomplete grasp of the semantic map's spatial topology, this action fundamentally contradicts the correct trajectory. The multidimensional action verifier evaluates this proposal and promptly rejects it. The verifier identifies that flying northwest not only fails to advance the navigation task but also induces unnecessary detouring (i.e., overlapping with the historical flight trajectory).

Triggered by this rejection, the system initiates the verification-triggered multimodal retriever. By synthesizing the environmental description and the verifier's analytical rationale, the system generates a precise query to retrieve relevant external knowledge. The retrieved knowledge consists of a successful historical navigation case that explicitly delineates the reasoning process behind directional selection. This success case, alongside the error analysis, is subsequently fed into the hierarchical reasoner for the second iteration ($k=2$). Augmented by this reasoning knowledge, the reasoner successfully corrects its output, predicting ``east'' as the optimal direction. The multidimensional action verifier subsequently confirms that moving east effectively advances the task without causing trajectory deviation, accepting it as the final action for this time step.

As observed in Step 5, the eastward flight effectively brings the UAV closer to the stage goal (denoted by the yellow bounding box in $o_5$). By contrast, an open-loop reasoning paradigm would have blindly executed the flawed ``northwest'' decision generated in the first iteration. Such an action will yield no progress toward the sub-instruction and actively distance the UAV from the final trajectory destination (red bounding box). This comparative analysis explicitly highlights how correct advancement toward stage goals directly facilitates the discovery of final objectives, thereby demonstrating that the closed-loop reasoning paradigm exhibits superior decision-making reliability compared to the open-loop paradigm. 

It is worth emphasizing that the retrieved successful case suggests moving "southeast," while the hierarchical reasoner ultimately outputs "east." This discrepancy indicates that the reasoner does not simply replicate the case knowledge; instead, it autonomously analyzes and judges based on the reasoning rationale provided by the supplemented case, thereby correctly deriving the optimal direction. This not only demonstrates the strong inferential capability of the hierarchical reasoner but also validates that precise and appropriate knowledge supplementation can effectively unlock the intrinsic reasoning potential of MLLMs.

\subsubsection{Failure Case: Verification-Induced False Positive}
While the proposed closed-loop paradigm substantially enhances navigation robustness, its efficacy heavily hinges on the multidimensional action verifier's capacity to precisely discern whether a proposed action contributes positively to task progression. When encountering landmarks with intricate geometric contours or those situated at the boundaries of the semantic map, the verifier is prone to judgment errors, leading to severe false positives (i.e., erroneous acceptance). As illustrated in Figure~\ref{fig_7}, given the sub-instruction ``Navigate near the One Stop Building to establish proximity to the landmark,'' the hierarchical reasoner generates a ``northeast'' action at Step~4 based on the instruction and the semantic map ($m_5$) to approach the target. However, due to the highly complex geometric layout of the One Stop Building and its peripheral placement on the map grid, the verifier misjudges the landmark's actual boundary coordinates, leading it to erroneously accept the ``northeast'' decision. This verification failure allows the UAV to traverse right through the optimal subtask termination zone without stopping, thereby squandering the critical window for task completion.

A comparison between \( o_4 \) and \( o_5 \) reveals that the agent had already successfully entered the destination area but failed to initiate a halt. Such a phenomenon introduces a severe divergence in evaluation metrics: it yields a deceptively high OSR but a drastically compromised SR and path efficiency. Furthermore, improper or delayed termination of a localized subtask invariably propagates spatial compounding errors, compounding the operational difficulty of executing subsequent navigation instructions, and ultimately leading to a high NE.

To mitigate such localized failure modes, further optimization of the verifier's performance can be pursued. First, after the initial round of SFT, the model still exhibits certain suboptimal predictions. A second round of minor incremental fine-tuning can be conducted using the original training dataset without introducing additional annotated samples. This training phase adopts a lower learning rate and fewer epochs compared to the first round, continuing to converge on the original data distribution to better fit the feature patterns of hard-to-converge samples and rectify remaining prediction deficiencies. Meanwhile, a portion of general samples is mixed in to constrain the magnitude of parameter updates, thereby avoiding overfitting and preventing degradation of its original capabilities. Second, although the current verifier is aligned via supervised SFT and token-level GRPO, future training paradigms could transition to Group Sequence Policy Optimization (GSPO) to inherently stabilize long-sequence policy updates at the sequence level. By combining GSPO with fine-grained reward functions that penalize intermediate logical fallacies in the judgment rationale, the verifier can be incentivized to generate more rigorous spatial justifications and avoid boundary misjudgments without incurring prohibitive training costs.

\section{Conclusion}

In this paper, we propose CLOSER-VLN, a closed-loop reasoning mechanism that enables on-demand dynamic supplementation of external information. The system iteratively executes the cycle of decision generation, decision verification and evaluation, and navigation knowledge retrieval until the self-verified commentator determines that the current decision is sufficiently well-grounded or the maximum number of iterations is reached. Extensive experiments on the CityNav dataset demonstrate that the CLOSER-VLN not only achieves substantial performance gains over state-of-the-art baselines but also exhibits stronger generalization ability to unseen environments.

Future work will focus on: (1) distilling lightweight models for deployment to minimize edge-side latency; (2) integrating a multimodal world model to enhance the commentator's logical reasoning ability while reducing its training cost; and (3) implementing agent-based learning mechanisms to enable the exemplar database to continuously integrate successful cases for dynamic knowledge accumulation and self-evolution.


\bibliographystyle{IEEEtran} 
\bibliography{ref}    

\end{document}